\documentclass{article}

\usepackage{amsmath}
\usepackage{algorithm}
\usepackage{algpseudocode}
\usepackage{graphicx}
\usepackage{subcaption}
\usepackage{booktabs}
\usepackage{multirow}
\usepackage{amssymb}
\usepackage[final]{corl_2026} 
\usepackage{tcolorbox}
\tcbuselibrary{breakable, skins}
\newtcolorbox{promptbox}[1]{                                                                                                                                                                   
      colback=gray!5!white,                                                                                                                                                                      
      colframe=gray!75!black,                                                                                                                                                                    
      title=#1,                                                                                                                                                                                  
      fonttitle=\bfseries,                                                                                                                                                                       
      boxrule=0.5mm,                                                                                                                                                                             
      arc=2mm,                                                                                                                                                                                   
      left=5pt, right=5pt, top=5pt, bottom=5pt,                                                                                                                                                  
      breakable,   
  }
  
\title{Bayesian Active Learning for Intent Disambiguation in Interactive Robot Planning}
\author{
  Huao Li\\
  Massachusetts Institute of Technology\\
  \texttt{huaol@mit.edu} \\
  \And
    Carson Sobolewski\\
  Massachusetts Institute of Technology\\
  \texttt{csobo@mit.edu} \\
  \And
   Augustinos Saravanos\\
  Massachusetts Institute of Technology\\
  \texttt{asaravan@mit.edu} \\
  \And
    William Tan\\
  GE Vernova\\
  \texttt{yewteck.tan@gevernova.com} \\
    \And
    John Karigiannis\\
  GE Vernova\\
  \texttt{John.Karigiannis@gevernova.com} \\
    \And
    Chuchu Fan\\
  Massachusetts Institute of Technology\\
  \texttt{chuchu@mit.edu} \\
}

\begin{document}
\maketitle

\begin{abstract}
    Interactive robot planning requires robots to infer and execute human intentions from natural language instructions that are often ambiguous, incomplete, or underspecified. Although large language models (LLMs) provide a powerful interface for clarification, relying on the generative model to drive a multi-turn conversation can introduce systematic failures. We propose a Bayesian framework~\footnote{Project website: \href{https://www.huao-li.com/bal/}{https://www.huao-li.com/bal/}} that treats clarification as an active learning problem over grounded Signal Temporal Logic (STL) task specifications. Our method uses LLMs to initialize candidate formal specifications and translate informative contrasts into natural-language clarification questions, while Bayesian optimization maintains uncertainty estimates over user intent and selects queries that maximize information gain. Because queries are selected in a learned STL embedding space, the system can also ask about plausible specifications that the LLM did not propose. After convergence, the inferred STL specification is passed to a formal planner to synthesize a verifiable robot trajectory. Across four simulated and real-world task domains, our approach generally achieves higher task satisfaction and requires fewer clarification rounds than LLM baselines, while helping smaller models close the performance gap against larger reasoning models.
\end{abstract}


\section{Introduction}

\begin{figure*}
    \centering
    \includegraphics[width=\linewidth]{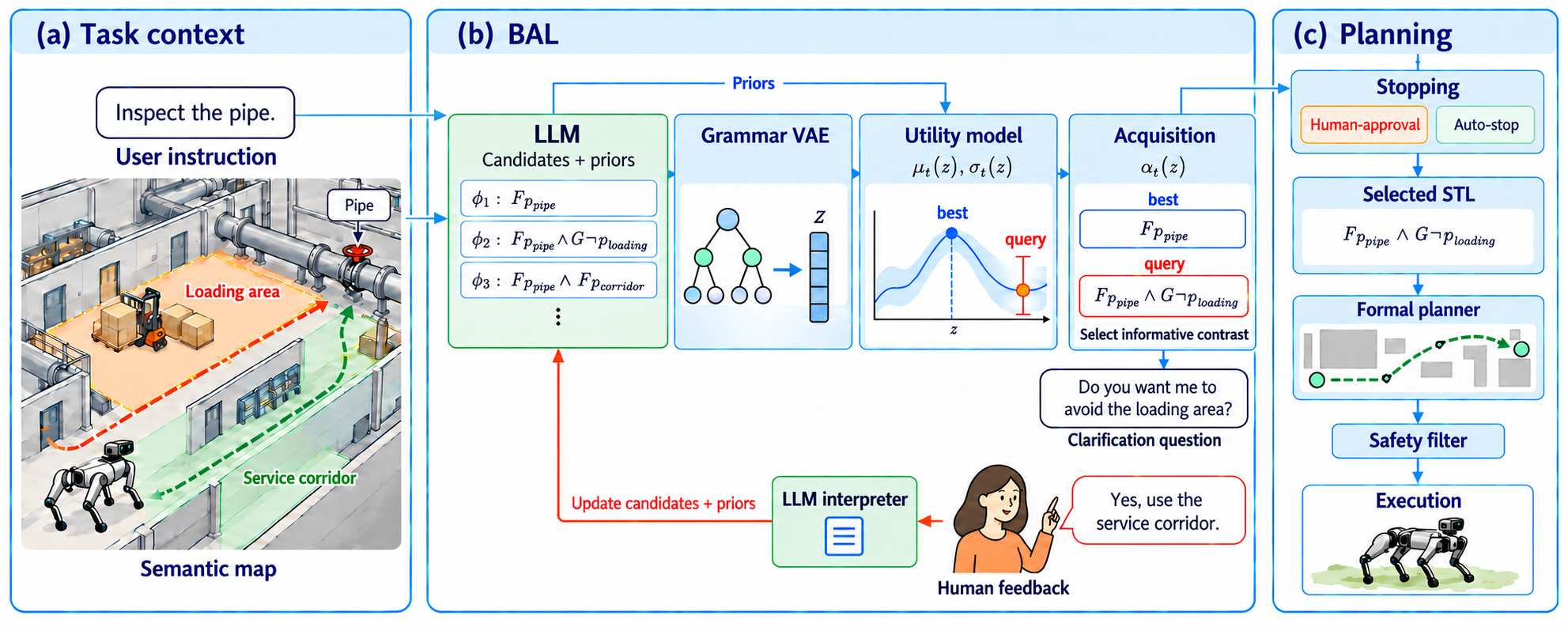}
    \caption{
Overview of the proposed Bayesian active learning framework.
An LLM proposes candidate specifications with priors, and a grammar VAE embeds them in a latent space where a Gaussian Process (GP) models utility. Queries are sampled by the acquisition function and translated into clarification questions by LLMs. User answers update the GP and the candidates. The final converged specification is decoded and passed to a formal planner for robot execution.
}
    \label{fig:framework}
\end{figure*}
Interactive robot planning requires systems to convert natural language instructions from human users into verifiable plans that can be executed in unstructured environments. While Large Language Models (LLMs) provide a powerful natural language interface for this interaction, off-the-shelf LLMs struggle to reliably generate long-horizon plans or physically ground their plans within the complex spatio-temporal constraints of the real world \cite{ahn2022saycan,liang2023code}. Consequently, recent research has sought to combine the semantic reasoning capabilities of LLMs with the rigorous safety guarantees of formal solvers \cite{liu2023llmp,chen2024autotamp,liu2023lang2ltl,chen2023nl2tl,hao2025planning}. In these hybrid architectures, the LLM translates natural language instructions into formal task specifications (e.g., temporal logic), and a formal solver generates the low-level robot plan, ensuring safety and transparency.

Despite this progress, most language-grounded planning pipelines implicitly assume that the user's initial instruction is sufficiently complete. However, in real-world human-robot interaction, users often rely on shared context, use ambiguous references, or leave safety and social constraints unstated. For instance, the user may provide a vague instruction such as ``Go to Tom's office,'' which lacks critical physical grounding (e.g., where the office is located), commonsense constraints (e.g., slowing down when passing near humans), and user preferences (e.g., which Tom, by what time). While some approaches enable LLMs to ask clarification questions when uncertain \cite{ren2023robotsask, hwang2025masked}, mounting evidence indicates that state-of-the-art LLMs suffer from context loss in multi-turn dialogues \cite{laban2025lost} and frequently exhibit sycophancy or over-confidence when interacting with users \cite{groot2024overconfidence}. These issues motivate a robot planning system that treats clarification as an active information-gathering problem with theoretical grounding.

Human-robot interaction has a long history of learning user preferences from demonstrations, comparisons, corrections, and language feedback. Bayesian preference learning, inverse reinforcement learning, and learning-from-demonstration methods can actively query users to infer reward functions based on task-relevant features~\cite{sadigh2017active, bajcsy2017learning, jeon2020reward}. However, many such methods assume that preferences are linear combinations of predefined features, such as trajectory length, smoothness, or distance to humans. Recent work uses language models to construct richer task abstractions or infer latent preferences from behavior \cite{peng2024algae,peng2024plga}, but the learned preference is still often represented as a reward or feature abstraction whose connection to formal task satisfaction is indirect. In safety-critical robot planning, this creates a gap between what the system learns from the human and what the downstream planner can formally verify.

We address this gap by modeling the user's intended preference as a formal task specification grounded in the robot's environment. Specifically, we represent candidate intents as Signal Temporal Logic (STL) formulas over atomic propositions grounded in the task scenario. STL then provides a structured hypothesis space for representing temporally extended navigation and manipulation preferences, including goals, avoidance constraints, ordering constraints, and timing requirements. Instead of assuming a single translation from language to logic, our system maintains uncertainty over possible STL specifications and actively asks the user questions that reduce this uncertainty via Bayesian active learning. Within this framework, LLMs serve as language-based support modules: they initialize the preference model by proposing and scoring candidate task specifications from the initial instruction and mediate the natural language conversation with human users. After the active learning loop converges, the system selects the most likely user-preferred STL formula and passes it to a formal planner, such as an STL-constrained mixed-integer planner \cite{sun2022multi}, to synthesize an optimal, collision-free trajectory.

Our contributions are threefold. First, we formulate clarification in language-conditioned robot planning as Bayesian active learning over formal task specifications. This casts human-robot dialogue as an information-gathering process, where each question is selected to maximize expected mutual information about the user's intent. Because queries are sampled directly in a learned STL embedding space, the robot can also challenge the user with specifications that the LLM never proposed. Empirically, this formulation recovers user intent with fewer clarification rounds than LLM-driven dialogue alone. Second, we model user intent as grounded STL specifications rather than unconstrained language plans or implicit rewards. This enables the inferred intent to be verified by a formal planner, while preserving the compositionality and interpretability needed for real-world deployment. Third, our framework is flexible across models, domains, and planning backends. We evaluate it with six LLMs across two simulated and two real-world environments spanning navigation and manipulation, and show that the same framework can incorporate different context inputs, intermediate task representations, and downstream solvers.

\section{Related Work}

\subsection{Language-Conditioned Formal Planning}
Recent work integrates LLMs with symbolic or formal planners to combine flexible language understanding with verifiable plan generation. LLM+P translates natural-language instructions into PDDL problems for classical planning \cite{liu2023llmp}, while related systems map language into temporal-logic specifications or task-and-motion planning constraints \cite{chen2024autotamp,chen2023nl2tl,liu2023lang2ltl,liu2024lang2ltl2}. Other approaches use foundation models as high-level formalizers that generate planning problems, domain rules, or optimization formulations from natural language \cite{hao2025simulation,hao2025planning}. These methods show that LLMs can help bridge informal language and formal solvers, but they primarily focus on improving the translation from a given instruction to a formal representation. Our work instead focuses on the preceding interaction problem: when the initial instruction is underspecified, the robot must actively infer which formal specification the user intends.

\subsection{Multi-Turn Clarification}

When human instructions are ambiguous or underspecified, the robot must elicit clarification. However, resolving complex specifications and safety constraints purely through multi-turn dialogue driven by LLMs poses structural risks. LLMs can get lost in multi-turn conversations by making incorrect assumptions and then over-relying on them in later turns \cite{laban2025lost}. Additionally, LLMs frequently exhibit sycophancy or over-confidence when interacting with users \cite{groot2024overconfidence}. Therefore, researchers attempt to guide clarification with external verifiers for more robust correction signals \cite{stechly2025self}. For example, KnowNo utilizes conformal prediction to quantify LLM uncertainty and proactively query human users when confidence is low \cite{ren2023robotsask}. More recently, Masked IRL grounds clarification in observed demonstrations, using LLM reasoning to resolve ambiguous instructions before translating the clarified intent into reward-relevant state masks \cite{hwang2025masked}. Our framework treats disambiguation as a Bayesian active learning problem over formal hypotheses rather than relying purely on language-model memory. We leverage LLMs strictly for semantic support, offloading rigorous uncertainty estimation and query generation to the probabilistic model.

\subsection{Preference Learning in Human-Robot Interaction}

Human-robot interaction has a long history of learning latent user preferences from demonstrations, corrections, comparisons, and language feedback. Active preference learning methods infer reward functions by querying users about informative trajectory comparisons or demonstrations \cite{sadigh2017active,bajcsy2017learning,jeon2020reward}. More recent language-guided approaches use LLMs to extract interpretable features, propose abstractions, or condition preference learning on natural-language feedback \cite{peng2024algae,peng2024plga,mahmud2025maple}. Related interactive planning methods also use language feedback to discover predicates and operators for symbolic planning \cite{han2024interpret}. These methods improve sample efficiency and interpretability, but the learned object is often a reward function, feature abstraction, or predicate set whose connection to formal task satisfaction is indirect. Our framework instead places Bayesian active learning directly over grounded STL specifications, integrating sample-efficient human feedback with a formal representation that can be verified by the downstream planner.

\section{Method}

We formulate interactive robot clarification as a \emph{Bayesian active learning} (BAL) problem, as shown in Fig.~\ref{fig:framework}. Given an ambiguous natural-language instruction, the robot must infer the formal task specification that best represents the user's latent intent. Specifically, our method infers a grounded Signal Temporal Logic (STL) specification and then uses a formal planner to synthesize a trajectory satisfying that specification.

\subsection{Problem Formulation}

Let $\mathcal{G}$ denote a grounded robot environment containing objects, regions, semantic labels, and geometric constraints. The environment induces a set of grounded atomic propositions $\Pi_{\mathcal{G}}$, such as whether the robot is inside a target region or avoids an unsafe area. A robot trajectory is denoted by $\xi=(x_0,\ldots,x_H)$, where $x_t$ is the robot state at time $t$.

The user provides an initial instruction $I_0$, which may be ambiguous or incomplete. We assume that the user's intent can be represented by a formal task specification $\phi \in \Phi_{\mathcal{G}}$, where $\Phi_{\mathcal{G}}$ is the set of valid specifications over the grounded propositions in $\mathcal{G}$. The robot's objective is to infer the intended task specification and synthesize a trajectory satisfying it.

We introduce a utility function $U^\star:\Phi_{\mathcal{G}}\rightarrow\mathbb{R}$, where $U^\star(\phi)$ measures how well candidate specification $\phi$ matches the user's latent intent. The inference objective is
\begin{equation}
    \phi^\star \in \arg\max_{\phi\in\Phi_{\mathcal{G}}} U^\star(\phi).
    \label{eq:spec-objective}
\end{equation}

\subsection{STL Specification Space}

We use Signal Temporal Logic (STL) to represent temporally extended task intent. STL is defined recursively: grounded predicates form atomic formulas, and more complex formulas are built by composing existing formulas with Boolean and temporal operators. This recursive structure makes task specifications modular and interpretable, since each subformula corresponds to a meaningful constraint such as reaching a goal, avoiding an unsafe region, or satisfying an ordering relation. Formally, STL formulas are constructed from grounded predicates, Boolean operators, and temporal operators such as eventually $\mathbf{F}_{[a,b]}$, always $\mathbf{G}_{[a,b]}$, and until $\mathbf{U}_{[a,b]}$. For example, $\mathbf{F}_{[0,30]}p_{\mathrm{goal}}$ requires the robot to reach the goal within 30 seconds, while $\mathbf{G}_{[0,H]}\neg p_{\mathrm{unsafe}}$ requires the robot to avoid unsafe regions throughout the task horizon. We write $\xi\models\phi$ if trajectory $\xi$ satisfies formula $\phi$. STL also provides a robustness score $\rho^\phi(\xi)$, where $\rho^\phi(\xi)\geq 0$ implies satisfaction. This allows the final inferred specification to be passed directly to a formal trajectory optimizer.
\subsection{Latent Utility Model over Specifications}

Because $\Phi_{\mathcal{G}}$ is discrete and combinatorial, direct search over formulas is difficult. We therefore embed STL formulas into a continuous latent space using a grammar VAE~\citep{kusner2017grammar}. Let $E_\psi(\phi)=z$ encode a formula into $z\in\mathcal{Z}$, and let $D_\psi(z)=\phi$ decode a latent point into a syntactically valid STL formula. We then define the latent utility function as $u^\star(z)=U^\star(D_\psi(z))$.

The framework maintains a probabilistic \emph{latent utility model} over $u^\star$, which can be any surrogate that provides a posterior mean and uncertainty. In our implementation, we use a Gaussian process (GP). Given observations $\mathcal{D}_t=\{(z_i,y_i)\}_{i=1}^{n_t}$, where $y_i$ is a direct utility estimate for the candidate specification $D_\psi(z_i)$, the GP posterior is
\begin{equation}
    u(z)\mid \mathcal{D}_t
    \sim
    \mathcal{N}\bigl(\mu_t(z),\sigma_t^2(z)\bigr).
    \label{eq:gp-posterior}
\end{equation}

The posterior mean $\mu_t(z)$ estimates how well a specification matches the user's intent, while the posterior variance $\sigma_t^2(z)$ captures uncertainty. The LLM initializes the model by proposing a candidate set $\mathcal{C}_0$ with prior utility estimates based on the initial instruction and task context, thereby guiding the initial search toward semantically plausible regions of the specification space. We write $\mathcal{Z}_{\mathcal{C}_t}=\{E_\psi(\phi):\phi\in\mathcal{C}_t\}$ for the encoded candidates at round $t$.

\subsection{Active Clarification Query Selection}

At clarification round $t$, the system selects two specifications
to contrast. The first is the current best estimate,
$z_t^{\text{best}} =
\arg\max_{z\in\mathcal{Z}_{\mathcal{C}_t}}\mu_t(z)$.
The second is an informative challenger selected by an acquisition
function that combines expected information gain with a utility bonus:
\begin{equation}
    z_t^{\text{query}}
    =
    \arg\max_{z\in\mathcal{Z}_t}\alpha_t(z),
    \qquad
    \alpha_t(z)
    =
    I(y;u\mid z,\mathcal{D}_t)+\lambda\mu_t(z).
    \label{eq:acquisition}
\end{equation}
Here, $y$ denotes a noisy utility observation, and
$I(y;u\mid z,\mathcal{D}_t)$ measures the expected information
gained about the latent utility function by querying $z$.
The parameter $\lambda\geq 0$ controls the preference for
high-utility challengers. In our experiments, we set $\lambda=0$
to select challengers purely by information gain, while the
current best specification provides the exploitative anchor.

The search domain $\mathcal{Z}_t\subseteq\mathcal{Z}$ extends
beyond the encoded LLM candidates. In our implementation, we
estimate mutual information using Monte Carlo samples and
optimize the acquisition function continuously over bounded
latent coordinates. The resulting challenger is decoded through
the grammar VAE and added to the candidate pool. This allows BAL
to explore specifications that the LLM has not proposed, rather
than merely re-ranking its candidates. We decode the selected
points as
$\phi_t^{\text{best}}=D_\psi(z_t^{\text{best}})$ and
$\phi_t^{\text{query}}=D_\psi(z_t^{\text{query}})$.

The LLM translates the contrast between these specifications
into a natural-language clarification question $Q_t$ targeting
a distinguishing attribute, such as object identity, timing,
ordering, or avoidance. For example, specifications that differ
in a target's arrival window can prompt a question about the
desired timing constraint.

After receiving the user's free-form answer $A_t$, the LLM
interprets it in the context of the question and contrasting
specifications, updates the candidate pool, and assigns fresh
utility scores to affected candidates. These scores are treated
as noisy observations of $u^\star$ and appended to
$\mathcal{D}_t$. Candidate generation is conditioned on the
original instruction and accumulated dialogue, and a
deduplicated cumulative pool retains previously proposed
specifications. The utility model is refit before the next
round. Appendix~\ref{app:workflow} provides implementation
details.

\subsection{Stopping, Specification Selection, and Planning}
\label{sec:stop}

At each round, the current estimate $\phi_t^{\text{best}}$ is passed to a formal planner, which synthesizes a feasible trajectory by solving
\begin{equation}
    \xi_t
    =
    \arg\min_{\xi\in\Xi_{\mathrm{feasible}}} C(\xi)
    \quad
    \mathrm{s.t.}\quad
    \xi \models \phi_t^{\text{best}}.
    \label{eq:planning}
\end{equation}
Clarification ends when a generic stopping test $\textsc{Stop}(\phi_t^{\text{best}},\xi_t,\mathcal{D}_t)$ succeeds or the query budget $B$ is exhausted, and the system returns $\hat{\phi}=\phi_t^{\text{best}}$ and $\xi^\star=\xi_t$ (See Algorithm~\ref{alg:lgbal} in Appendix). We keep $\textsc{Stop}$ generic; for example, it can succeed when the user approves the proposed trajectory $\xi_t$. This design separates intent inference from trajectory optimization: the utility model infers the intended STL specification, while the planner generates a feasible trajectory satisfying it.

\section{Experiments}
\label{sec:exp}
We evaluate BAL in two simulated domains and two real-world deployments. The simulated experiments test whether BAL recovers user intent from ambiguous instructions more efficiently than LLM-based clarification. The real-world experiments test deployment on a Unitree Go2 quadruped.

\begin{figure*}[t]
    \centering
    \begin{subfigure}[b]{0.23\textwidth}
        \centering
        \includegraphics[width=\textwidth]{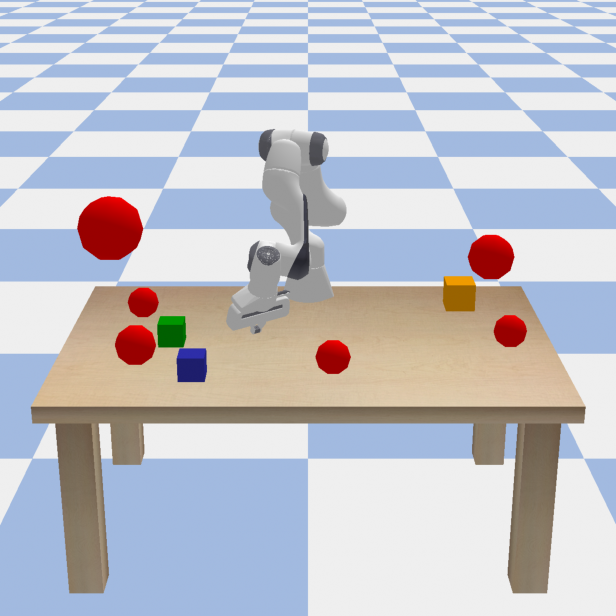}
        \caption{Franka Panda}
        \label{fig:env_panda}
    \end{subfigure}
    \hfill
    \begin{subfigure}[b]{0.23\textwidth}
        \centering
        \includegraphics[width=\textwidth]{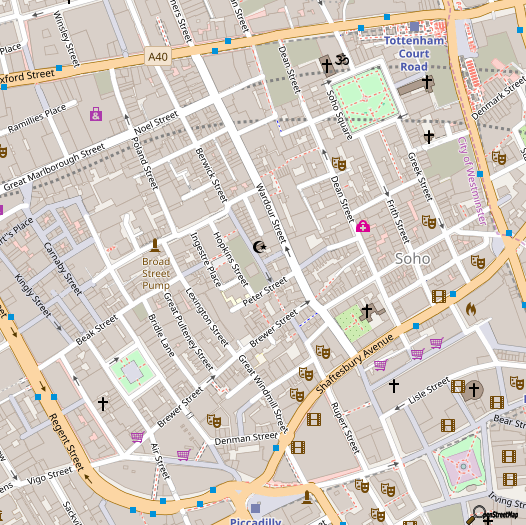}
        \caption{City}
        \label{fig:env_city}
    \end{subfigure}
    \hfill
    \begin{subfigure}[b]{0.23\textwidth}
        \centering
        \includegraphics[width=\textwidth]{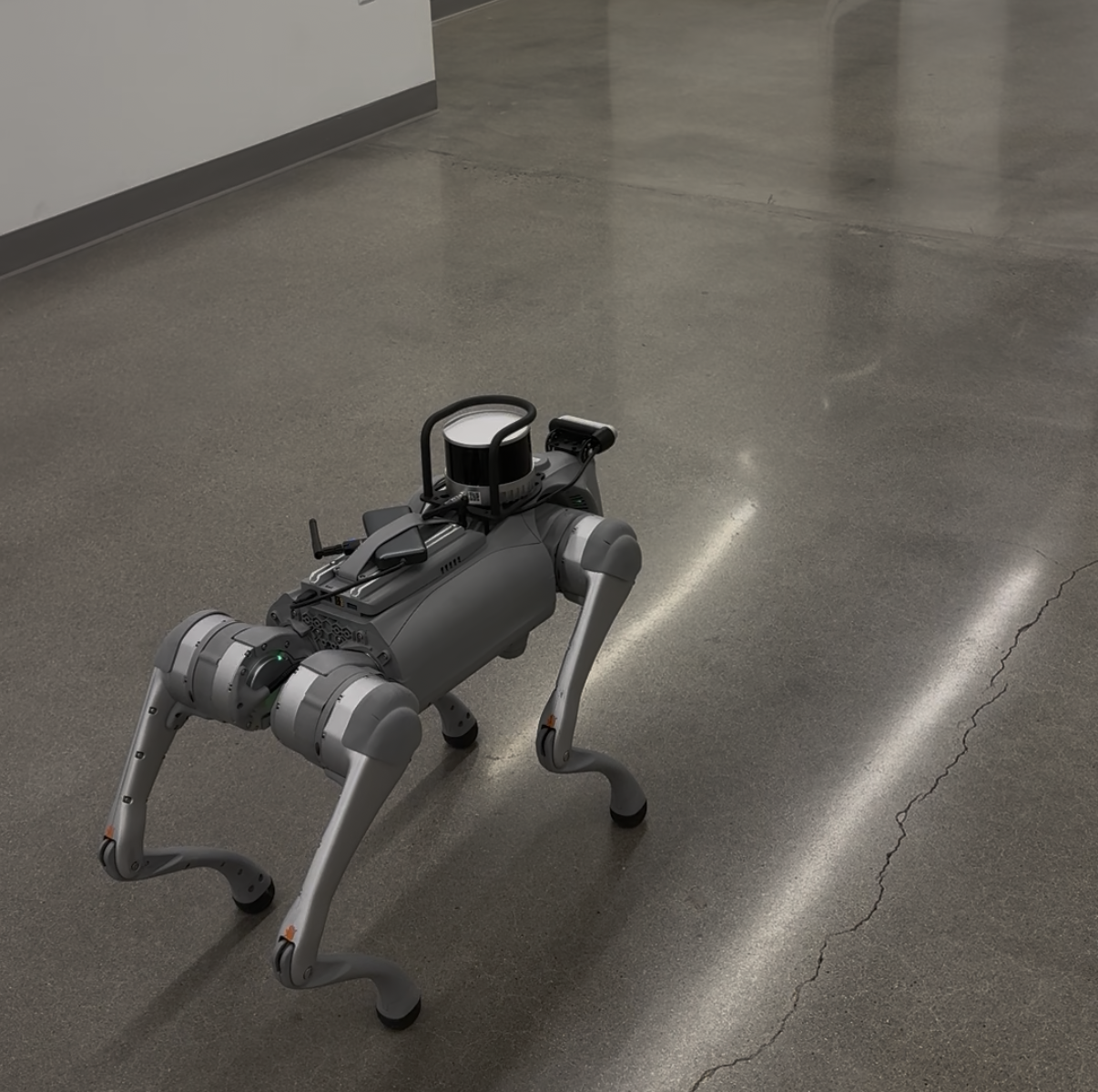}
        \caption{School}
        \label{fig:env_school}
    \end{subfigure}
    \hfill
    \begin{subfigure}[b]{0.23\textwidth}
        \centering
        \includegraphics[width=\textwidth]{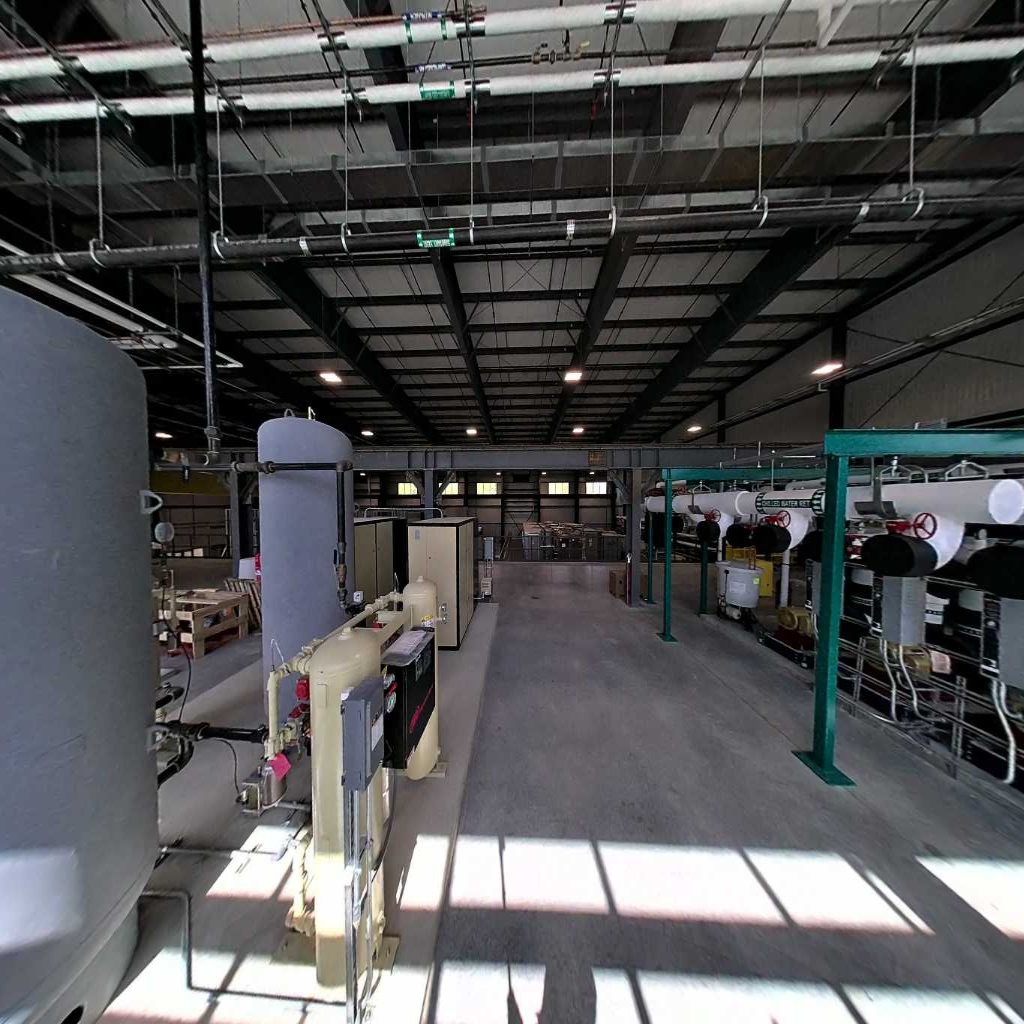}
        \caption{Factory}
        \label{fig:env_factory}
    \end{subfigure}
    \caption{Benchmark environments used in simulation and real-world deployment.}
    \label{fig:envs}
\end{figure*}

\subsection{Simulation Domains}

\paragraph{Franka Panda}
We use a 7-DoF Franka Panda tabletop manipulation domain~\citep{gaz2019dynamic}. The robot must reach specified colored objects while avoiding unsafe objects. Each scene contains 14 objects with semantic labels, colors, shapes, and metric positions. The LLM receives the robot pose, workspace bounds, object attributes, and admissible predicates. STL specifications are defined over object-level propositions such as \texttt{reach(orange\_cup)} and \texttt{avoid(red\_ball)}. We generate trajectories using a gradient-based optimizer that maximizes STL robustness.


\paragraph{City}
We use a 2D city navigation domain based on OpenStreetMap environments from Lang2LTL~\citep{liu2023lang2ltl,liu2024lang2ltl2}. Each map contains 40 landmarks with possible grounding ambiguity, such as repeated amenity types or similar street names. The environment is represented as a finite semantic transition system, and tasks are expressed in Linear Temporal Logic (LTL). Each inferred LTL formula is translated into a B\"uchi automaton and solved on the product system~\citep{baier2008principles}. Note although our framework is defined over STL, LTL can be viewed as a restricted Boolean-time fragment of STL without explicit metric time intervals, so this benchmark provides a valid evaluation setting for our specification inference and clarification pipeline.

\subsection{Test Scenarios}

Tasks are generated from four temporal-logic template families following prior STL benchmark design~\cite{meng2025telograf}: \emph{single-goal} tasks require reaching one target, \emph{multi-goal} tasks require satisfying multiple goals, \emph{sequential} tasks impose an ordering over goals, and \emph{partial-order} tasks specify precedence constraints without requiring a total ordering. For each domain, we randomly generate ground-truth specifications $\phi^\star$ across these templates. Each specification contains between 2 and 12 grounded objects or regions, including goals, obstacles, and temporal dependencies. Time intervals are sampled randomly for STL tasks. We filter all generated specifications through the corresponding planner to ensure that each benchmark instance is feasible.

For every ground-truth specification, we generate an ambiguous natural-language instruction by hiding the grounding or temporal information. The ambiguity generation procedure is designed to cover three common sources of ambiguity: \emph{referential ambiguity}, where multiple objects or landmarks share similar attributes; \emph{commonsense or safety ambiguity}, where task-relevant constraints are omitted; and \emph{preference ambiguity}, where the user leaves goals, obstacles, or priorities unspecified~\cite{ivanova2025ambik}. In simulation, the human user is modeled by an LLM proxy conditioned on $\phi^\star$. The proxy generates the initial ambiguous instruction and answers subsequent clarification questions. For each test case, the proxy and initial instruction are fixed across methods, ensuring that all methods receive the same interaction context.

To study information-seeking in human-robot dialogue, we assume low-initiative users: the initial instruction omits some information, and each subsequent response clarifies only one aspect of the intended task. This information bottleneck prevents the robot from relying on overly general, open-ended questions, thereby making clarification question selection critical. Both the LLM proxy and the human participants in Sec.~\ref{sec:user} are instructed to follow this protocol. Under this constrained, closed-ended setting, we expect LLM proxies and human users to exhibit similar interaction patterns.

\subsection{Baselines}

We compare BAL against two LLM-based clarification baselines. All methods use the same base LLM, the same simulated user proxy, and the same context information in prompts. \textbf{LLM Clarification (LM).} The LM baseline asks free-form natural-language clarification questions and then directly outputs the final executable plan, such as an action sequence or waypoint trajectory, without producing an intermediate formal specification. \textbf{LLM Clarification + Planner (LM+P).} The LM+P baseline asks free-form clarification questions, then translates the clarified instruction into an STL specification. The resulting specification is passed to the same downstream planner used by BAL. This baseline isolates the benefit of explicit Bayesian uncertainty modeling from the benefit of using a formal planner.



\subsection{Metrics}
In the main experiments, we instantiate the stopping test in Algorithm~\ref{alg:lgbal} as human approval for all methods: each method may ask up to $B=10$ clarification questions, and the interaction stops early once the user approves the proposed trajectory, i.e., once the trajectory satisfies the ground-truth specification. Each method is evaluated by \emph{Success Rate}, whether the final executed trajectory satisfies the ground-truth specification, and \emph{Rounds}, the number of clarification turns required before first approval. \emph{Time} measures wall-clock runtime per task, including LLM calls, latent-space inference, Bayesian optimization, and planning. All results are reported as mean~$\pm$~standard error over all test cases.

  \subsection{Results}


\begin{table}[t]
    \centering
    \caption{Success rate (\%, $\uparrow$) and rounds of clarification ($\downarrow$)
    on Panda and City. Mean $\pm$ standard error.
    \textbf{Bold} marks the per-domain best.}
    \label{tab:main_results}

    \scriptsize
    \begin{tabular}{@{}lcccccc@{}}
        \toprule
        \multirow{2}{*}{Base Model}
        & \multicolumn{3}{c}{Panda} & \multicolumn{3}{c}{City} \\
        \cmidrule(lr){2-4} \cmidrule(lr){5-7}
        & BAL & LM & LM+P & BAL & LM & LM+P \\
        \midrule
        \multicolumn{7}{@{}l}{\textit{Success Rate (\%) $\uparrow$}} \\
        GPT-5.4               & \textbf{91.1$\pm$4.2} & 48.9$\pm$7.5 & 82.2$\pm$5.7 & \textbf{62.5$\pm$9.9}  & 29.2$\pm$9.3 & 20.8$\pm$8.3 \\
        GPT-5.4-mini          & \textbf{93.2$\pm$3.8} & 64.4$\pm$7.1 & 86.7$\pm$5.1 & \textbf{64.7$\pm$11.6} & 20.8$\pm$8.3 & 12.5$\pm$6.8 \\
        Claude-Opus-4-6       & \textbf{91.1$\pm$4.2} & 64.4$\pm$7.1 & 82.2$\pm$5.7 & \textbf{50.0$\pm$10.2} & 29.2$\pm$9.3 & 20.8$\pm$8.3 \\
        Claude-Sonnet-4-6     & \textbf{91.1$\pm$4.2} & 73.3$\pm$6.6 & 86.7$\pm$5.1 & \textbf{41.7$\pm$10.1} & 29.2$\pm$9.3 & 8.3$\pm$5.6  \\
        Gemini-3.1-Pro        & \textbf{91.1$\pm$4.2} & 82.2$\pm$5.7 & \textbf{91.1$\pm$4.2} & \textbf{54.2$\pm$10.2} & 33.3$\pm$9.6 & 25.0$\pm$8.8 \\
        Gemini-3.1-Flash-Lite & \textbf{93.3$\pm$3.7} & 64.4$\pm$7.1 & 91.1$\pm$4.2 & \textbf{66.7$\pm$9.6}  & 29.2$\pm$9.3 & 12.5$\pm$6.8 \\
        \midrule
        \multicolumn{7}{@{}l}{\textit{Rounds of Clarification $\downarrow$}} \\
        GPT-5.4               & \textbf{2.33$\pm$0.41} & 6.16$\pm$0.62 & 3.51$\pm$0.54 & \textbf{6.54$\pm$0.57} & 9.14$\pm$0.31 & 9.17$\pm$0.30 \\
        GPT-5.4-mini          & \textbf{2.09$\pm$0.36} & 5.40$\pm$0.58 & 3.24$\pm$0.51 & \textbf{6.46$\pm$0.70} & 9.42$\pm$0.30 & 9.53$\pm$0.28 \\
        Claude-Opus-4-6       & \textbf{2.93$\pm$0.47} & 4.69$\pm$0.62 & 3.38$\pm$0.52 & \textbf{7.25$\pm$0.53} & 8.75$\pm$0.39 & 8.94$\pm$0.41 \\
        Claude-Sonnet-4-6     & 3.09$\pm$0.48 & 4.49$\pm$0.60 & \textbf{3.02$\pm$0.50} & \textbf{8.11$\pm$0.49} & 8.67$\pm$0.39 & 9.39$\pm$0.33 \\
        Gemini-3.1-Pro        & \textbf{2.36$\pm$0.42} & 3.47$\pm$0.49 & 3.16$\pm$0.45 & \textbf{7.78$\pm$0.56} & 8.83$\pm$0.39 & 9.06$\pm$0.34 \\
        Gemini-3.1-Flash-Lite & \textbf{2.44$\pm$0.35} & 5.42$\pm$0.62 & 3.13$\pm$0.47 & \textbf{6.14$\pm$0.59} & 8.83$\pm$0.42 & 9.53$\pm$0.23 \\
        \bottomrule
    \end{tabular}
\end{table}

\begin{figure*}[h]
    \centering

    \begin{subfigure}[b]{0.31\textwidth}
        \centering
        \includegraphics[height=0.9\textwidth,trim={0 0 11.5cm 0},clip]{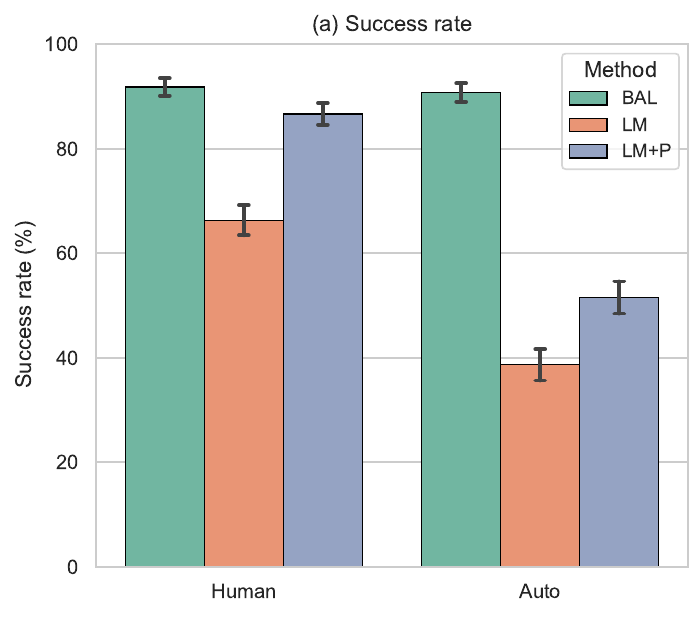}
        \caption{Success rate}
        \label{fig:auto_success}
    \end{subfigure}
    \hfill
        \begin{subfigure}[b]{0.31\textwidth}
        \centering
        \includegraphics[height=0.9\textwidth,trim={11.7cm 0 0 0},clip]{figs/fig_auto_vs_human_agg.pdf}
        \caption{Rounds of clarification}
        \label{fig:auto_rounds}
    \end{subfigure}
    \hfill
    \begin{subfigure}[b]{0.31\textwidth}
        \centering
        \includegraphics[height=0.9\textwidth]{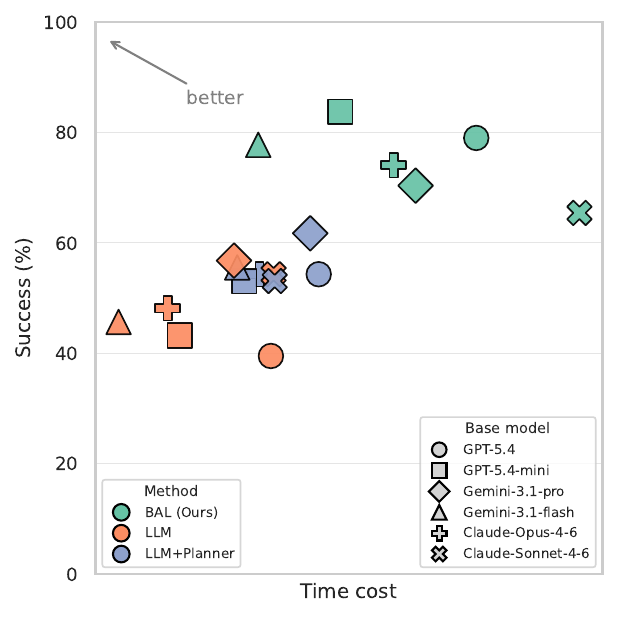}
        \caption{Cost performance analysis}
        \label{fig:pareto}
    \end{subfigure}

    \caption{\textbf{(a, b)} Comparison between human-approval and auto-stop modes on the Panda domain. BAL maintains high task success under auto-stop, while LLM-based baselines terminate earlier but suffer from substantially lower success rates.
    \textbf{(c)} Pareto scatter plot showing the trade-off between time cost and performance. BAL works best with smaller models near the top-left corner.}
    \label{fig:fig3}
\end{figure*}

\paragraph{Task Performance}

Table~\ref{tab:main_results} summarizes performance on Panda and City. Across all six base LLMs, BAL matches or exceeds the baselines in success rate and generally requires fewer clarification rounds. These results support the effectiveness of Bayesian intent modeling and active query selection across different language models.

Adding a formal planner improves the LM baseline on Panda, but LM+P underperforms LM on City. This suggests that planner augmentation alone is insufficient when the inferred specification does not capture the user's intent: the solver may produce a valid plan for an incorrect specification. BAL addresses this problem by maintaining uncertainty over candidate specifications and actively querying unresolved aspects of intent.

\paragraph{Cost Analysis}
Figure~\ref{fig:pareto} analyzes the trade-off between performance and computational cost. BAL introduces additional computation from latent-space inference and Bayesian optimization, but this overhead is often offset by more efficient clarification, whereas larger reasoning models rely on longer reasoning traces, tool calls, and repeated dialogue. Consequently, BAL with a smaller model can match or outperform direct clarification with a larger model at lower wall-clock cost, which is valuable for resource-constrained robot deployment. This suggests that BAL shifts much of the reasoning burden to its neuro-symbolic components: the LLM handles semantic interpretation and dialogue, while uncertainty estimation, query selection, and planning are handled by the probabilistic model and formal planner.

\section{Ablation Study: Stopping Criteria}

The main experiments use human-approval stopping, which requires the user to validate the proposed trajectory at each round. We additionally evaluate an \emph{auto-stop} mode on Panda, where the robot decides when clarification is complete and the user evaluates the final proposal afterward. BAL stops when the posterior differential entropy over the candidate pool fails to decrease for two consecutive rounds. The LLM baselines are prompted to stop when they judge the user's intent sufficiently clear.

As shown in Fig.~\ref{fig:auto_success}, BAL maintains similar success rates under auto-stop, whereas the LLM baselines show substantial declines. Clarification rounds must therefore be interpreted together with success (Fig.~\ref{fig:auto_rounds}). Under human-approval stopping, fewer rounds indicate faster recovery of a satisfying trajectory, subject to the query budget. Under auto-stop, fewer rounds can instead reflect premature termination. The LLM baselines exhibit both low success and low interaction counts, consistent with premature confidence in their intent estimates. BAL uses a few additional rounds for posterior uncertainty to stabilize while maintaining high success, suggesting that explicit uncertainty estimation provides a useful stopping signal.

\section{User Study}
  \label{sec:user}
We conducted a web-based user study to evaluate our BAL agent with real users in the City navigation domain. We recruited $N{=}30$ participants through Prolific after excluding incomplete submissions. Each participant completed six episodes with a maximum of five clarification rounds in each episode. We used a within-subject design with three methods (\textit{BAL}, \textit{LM}, and \textit{LM+P}), randomly assigned across episodes and counterbalanced using a Latin square. To facilitate efficient data collection, we reduced the task complexity to at most four targets and five clarification rounds per episode. All other settings matched the simulation experiments in Sec.~\ref{sec:exp}, except that real users replaced the LLM proxy. This study was approved by the IRB as an exempt research study. Table~\ref{tab:human_results} reports the per-participant means. A one-way repeated-measures ANOVA showed a significant main effect of method on success rate, $F(2,58){=}6.10$, $p{=}.004$, $\eta_p^2{=}.17$, and interaction rounds, $F(2,58){=}8.78$, $p{=}.001$, $\eta_p^2{=}.23$. \textit{BAL} achieved the highest success rate while requiring the fewest clarification rounds, supporting its effectiveness relative to the baselines in real-user interactions.

  \begin{table}[t]
  \centering
  \caption{User study results (mean $\pm$ SE over $N{=}30$ participants).}
  \label{tab:human_results}
  \begin{tabular}{lcc}
  \toprule
  Method & Success rate $\uparrow$ & Avg.\ rounds $\downarrow$ \\
  \midrule
  \textbf{BAL}  & $\mathbf{0.65 \pm 0.07}$ & $\mathbf{3.67 \pm 0.21}$ \\
  LM  & $0.48 \pm 0.07$          & $4.38 \pm 0.14$ \\
  LM+P & $0.37 \pm 0.06$          & $4.25 \pm 0.15$ \\
  \bottomrule
  \end{tabular}
  \end{table}



\section{Real-World Deployment}
  \label{sec:exp-real}


We deploy BAL in two physical sites: \textbf{School}, an academic indoor environment, and \textbf{Factory}, an industrial environment. In both sites, a Unitree Go2 quadruped executes the inferred specifications. We scan each environment with an XGrids PortalCam and build a metric-semantic scene representation containing object labels, 3D bounding boxes, and navigable regions. These elements define the grounded propositions used by BAL. The inferred specification is passed to a MILP planner, which produces waypoints satisfying the temporal-logic constraints. The Go2 tracks the waypoints using MPC, with a CBF-QP safety filter for local collision avoidance~\citep{ames2017control}. We run 10 ambiguous instructions at each site using Gemini-3.1-Flash-Lite and a query budget of $B=10$, with human-approval stopping. BAL achieves 90\% success rate on both domains with fewer round of clarification than baseline methods. More details are provided in the Appendix.

\section{Discussion}

Our results suggest that explicit uncertainty modeling can improve both clarification efficiency and stopping decisions in interactive robot planning. BAL generally achieves higher task satisfaction with fewer clarification rounds across models and domains. The auto-stop experiment further suggests that posterior uncertainty provides a useful stopping signal, reducing the premature termination observed in the LLM baselines. The runtime analysis shows that BAL with smaller models can match or outperform direct clarification with larger models, indicating that structured intent inference can reduce reliance on language-model scale. The user study supports the benefits of this approach beyond simulated-user interactions, while the physical deployments demonstrate its integration with robot planning and execution.

Several limitations remain. BAL assumes a known specification space and an accurate, relatively static environment representation; intent involving missing predicates or changing scene conditions may not be recovered. It also relies on LLMs for candidate generation, question verbalization, and response interpretation, whose errors can affect intent inference. Finally, our evaluation uses a constrained, low-initiative interaction protocol, and the human study covers one domain with 30 participants. Future work should investigate uncertain grounding and more varied human interactions, including whether users find the clarification process natural and useful.

\section*{ACKNOWLEDGMENT}

This work is supported by the MIT x GE Vernova Energy and Climate Alliance under the project of Autonomous Field Inspection with Physical AI: A Neural Certificate Guided Framework for Safe, Semantic-Aware Robotic Inspection.

\bibliography{example}

\newpage
\appendix
\section*{Appendix}

\section{Pseudocode of Bayesian Active Learning Algorithm}

\begin{algorithm}[h]
\caption{Language-Grounded Bayesian Active Learning}
\label{alg:lgbal}
\begin{algorithmic}[1]
\Require Instruction $I_0$, environment $\mathcal{G}$, budget $B$
\Ensure Specification $\hat{\phi}$, trajectory $\xi^\star$
\State $(\mathcal{C}_0,\mathcal{D}_0)\gets\Call{LLM-Initialize}{I_0,\mathcal{G}}$
\For{$t=0,\ldots,B$}
    \State Fit utility model $(\mu_t,\sigma_t^2)$ on $\mathcal{D}_t$
    \State $\phi_t^{\mathrm{best}}\gets D_\psi\big(\arg\max_{z\in\mathcal{Z}_{\mathcal{C}_t}}\mu_t(z)\big)$
    \State $\xi_t\gets\Call{STL-Plan}{\mathcal{G},\phi_t^{\mathrm{best}}}$
    \State \textbf{if} $\Call{Stop}{\phi_t^{\mathrm{best}},\xi_t,\mathcal{D}_t}$ \textbf{then break}
    \State $\phi_t^{\mathrm{query}}\gets D_\psi\big(\arg\max_{z\in\mathcal{Z}_t}\alpha_t(z)\big)$
        \Comment{Eq.~\eqref{eq:acquisition}}
    \State $Q_t\gets\Call{LLM-Question}{\phi_t^{\mathrm{best}},\phi_t^{\mathrm{query}}}$
    \State $A_t\gets\Call{QueryUser}{Q_t}$
    \State $(\mathcal{C}_{t+1},\mathcal{D}_{t+1})\gets\Call{LLM-Update}{A_t,\mathcal{C}_t,\mathcal{D}_t,\phi_t^{\mathrm{query}}}$
\EndFor
\State \Return $\hat{\phi}\gets\phi_t^{\mathrm{best}}$, $\xi^\star\gets\xi_t$
\end{algorithmic}
\end{algorithm}

\section{Additional Results}

\subsection{Real-world Deployment Results}
Table~\ref{tab:demo} summarizes the real-world deployment results. All approved tasks are executed successfully with zero collisions. These results show that the same BAL pipeline transfers to physical environments without heavy engineering effort.

  \begin{table}[h]
  \centering
  \caption{Real-world deployment results (mean $\pm$ SE over 10 runs per site).}
  \label{tab:demo}
  \setlength{\tabcolsep}{3pt}\footnotesize
  \begin{tabular}{@{}llccc@{}}
  \toprule
  Domain & Metric & BAL & LM & LM+P \\
  \midrule
  \multirow{2}{*}{School}  & Success rate (\%) $\uparrow$ & \textbf{90.0$\pm$9.5} & 60.0$\pm$15.5 & 80.0$\pm$12.6 \\
   & Avg.\ rounds $\downarrow$ & \textbf{3.50$\pm$1.13} & 6.30$\pm$1.17 & \textbf{3.50$\pm$1.16} \\
  \midrule
  \multirow{2}{*}{Factory} & Success rate (\%) $\uparrow$ & \textbf{90.0$\pm$9.5} & 80.0$\pm$12.6 & \textbf{90.0$\pm$9.5} \\
   & Avg.\ rounds $\downarrow$ & \textbf{2.90$\pm$0.90} & 4.30$\pm$1.27 & 4.10$\pm$1.04 \\
  \bottomrule
  \end{tabular}
  \end{table}

\section{Implementation Details}
\label{app:impl}

We provide additional implementation details for the three principal components
of our framework: (i) the per-domain context inputs that ground the candidate
specifications, (ii) the active learning workflow, and
(iii) the grammar VAE training pipeline. Unless stated otherwise, all values
below are the defaults used for the experiments reported in the main paper.

\subsection{Context Inputs Across Domains}
\label{app:context}

All four domains share a common context interface: each scene exposes (a) a set
of grounded propositions $\Pi_{\mathcal{G}}$ used as terminals in the STL/LTL
grammar, (b) a metric or topological description of those propositions for the
planner, and (c) a free-text rendering of the scene that is provided to the
LLM as part of the system prompt. The four domains differ only in the source
of these elements.

\paragraph{School}
The school environment is reconstructed offline from multi-camera LiDAR scans
using the SPARK Dynamic Scene Graph (DSG) library, yielding a hierarchical
graph with object, place, and room layers. Each scene-graph node is exported
with a unique node id, layer index, semantic label, and an axis-aligned 2D bounding box
$(x_{\min}, x_{\max}, y_{\min}, y_{\max})$. Bounding boxes smaller than
$1\,\text{m}\times 1\,\text{m}$ are padded to $1\,\text{m}$ to ensure
MILP-tractable footprints, and objects entirely outside any room polygon are
pruned. Rooms are linked to contained objects through DSG containment edges.
Each annotated area receives a human-readable description, a dominant color,
a MILP position vector $[b_0,b_1,b_2,b_3]$, and a parent-room pointer. The
LLM context lists rooms, contained objects, descriptions, the start pose in
real-world meters/seconds, and a fixed STL knowledge base describing the
admissible predicates.

\paragraph{City (Lang2LTL OpenStreetMap).}
The city domain reuses the Lang2LTL-2 OSM benchmark. Per-city landmark
catalogues are extracted from the benchmark dataset by aggregating every
landmark name that appears in the city map of a ground-truth
formula, slugifying it into an atomic-proposition identifier
(e.g.\ \texttt{dumpling\_cafe}), and inferring a coarse type tag
(``park'', ``hotel'', ``school'', \ldots) by keyword matching over the OSM
name. The resulting LLM context groups landmarks by type and lists each as
(id, type, description). Because LTL is purely topological, we represent the
environment as a finite semantic transition system rather than a metric scene
graph, and we explicitly do not provide spatial coordinates, grounding
ambiguity is the learning target.

\paragraph{Factory.}
The factory scene is reconstructed from multi-camera LiDAR scans based on the vendor's object segmentation and semantic annotation, with manual correction from domain experts. Each area record contains a MILP bounding box, a
parent room id, a visual category, and a structured \texttt{\_domain\_info}
block capturing equipment type, primary function and typical usage, mounting
type, material,
an inspection pose with view direction and standoff distance, and topology
edges to adjacent,upstream,downstream equipment (relationship type,
target id, free-text description). The LLM context surfaces the agent role
(``industrial inspection robot''), the room$\rightarrow$equipment tree, and
the same STL knowledge base used in the school domain.

\paragraph{Franka Panda.}
We created Tabletop scenes stored as JSON records containing the table z-height,
the x/y workspace bounds in the robot base frame, and a list of object
instances. Each object exposes an integer id, a name following a
\texttt{color\_type[\_index]} convention (e.g.\ \texttt{red\_block},
\texttt{purple\_cup\_2}), a 3D position in meters, an effective sphere
radius, and a role tag \texttt{goal} or \texttt{obstacle}. The LLM context
lists the table bounds and z-height, every object with its name, role,
position, and radius, and the semantic definitions of the manipulation
predicates: \texttt{reach($o$)} succeeds when the end-effector enters a
small sphere around object $o$ with the gripper oriented downward, and
\texttt{avoid($o$)} requires the end-effector to remain outside a buffered
sphere around $o$ throughout the trajectory.

For all four domains the same per-instance generation procedure is used to
build evaluation sets: a ground-truth STL/LTL formula is sampled from one of
the four template families (single-goal, multi-goal, sequential,
partial-order) with template-conditional bounds on the number of goals and
obstacles; the formula is filtered through the corresponding planner to
verify feasibility; an LLM is then prompted with the ground-truth formula and
the scene context to produce an ambiguous natural-language instruction by
hiding grounding, ordering, or timing information.

\subsection{Bayesian Active Learning Workflow}
\label{app:workflow}

\paragraph{Candidate pool.}
At every clarification round the workflow maintains two parallel tensors of
shape $(N_t, d)$: the raw VAE latents $\{z_i\}$ produced by encoding each
candidate STL formula, and the encoded latents $\{\tilde{z}_i\}$ produced by
the latent encoder that are actually consumed
by the GP. A cumulative pool of all candidates seen so far is also retained,
de-duplicated against the set of canonical STL string forms, so that
informative pairs proposed in earlier rounds remain reachable. Each round,
the LLM is prompted with the original instruction, the accumulated Q/A
history, and the scene context, and is asked to return a fresh batch of
$N_{\text{LLM}}=20$ candidate STL formulas together with prior utility
scores on a small integer scale.

\paragraph{Utility GP.}
We model the latent utility function $u^{\star}$ with a single-task
\texttt{SingleTaskGP} from BoTorch, wrapping a \texttt{ScaleKernel} over a
Matern-$3/2$ base kernel with per-dimension (ARD) lengthscales. Inputs are
normalized to the unit cube via a \texttt{Normalize} transform and outputs
are standardized via \texttt{Standardize}. The likelihood noise variance is
fitted jointly with the lengthscales by maximizing the exact marginal
log-likelihood through \texttt{fit\_gpytorch\_mll}; if the fit diverges we
fall back to a noise variance of $10^{-2}$ and refit. Each utility
observation corresponds to one LLM-assigned integer score for a decoded
candidate, treated as a noisy observation of $u^{\star}$. The GP is refit
from scratch after every clarification round.

The GP exposes (i) posterior mean and variance at arbitrary $\tilde{z}$,
(ii) the argmax of the posterior mean over the current pool (used to choose
the running ``best'' candidate $\tilde{z}^{\text{best}}$), and
(iii) analytic Expected Improvement for diagnostic purposes (not used as the
production acquisition).

\paragraph{Acquisition function.}
The acquisition function combines a Monte Carlo estimate of the mutual
information between the user's response and the latent utility with a
mean-bias term:
\[
\alpha_t(\tilde{z}) = w_{\text{MI}}\,\widehat{I}(y; u \mid \tilde{z}, \mathcal{D}_t)
\;+\; w_{\mu}\,\mu_t(\tilde{z}),
\]
where $\widehat{I}$ is estimated from $S_f=128$ posterior samples of $u$
and $S_y=128$ likelihood samples per posterior draw. We use weights
$(w_{\text{MI}}, w_{\mu})=(1.0, 0.0)$ for our reported experiments, so the
challenger is selected purely by information gain, while the exploit
candidate is supplied separately by the GP mean. Setting $w_{\mu}>0$
recovers the trade-off formulation of the main
paper.

\paragraph{Best vs.\ challenger selection.}
Each round we present the user with a pair $(\phi_t^{\text{best}},
\phi_t^{\text{query}})$. The best candidate is always the pool member with
the highest posterior mean, $\tilde{z}_t^{\text{best}} =
\arg\max_i \mu_t(\tilde{z}_i)$. We pose a continuous optimization problem,
fixing the best as an anchor and optimizing the challenger
$\tilde{z}_t^{\text{query}}$ over the encoder bounds with BoTorch's
\texttt{optimize\_acqf} ($q=1$, \texttt{num\_restarts}$=5$,
\texttt{raw\_samples}$=256$). The resulting latent point is then decoded
through the grammar VAE to recover an STL formula, which is appended to the
pool. This mode lets the system propose specifications that no LLM batch
had yet suggested.

\paragraph{Prior bootstrap.}
At round $t=0$ the GP is initialized from the LLM prior. The LLM is asked
to return integer utility scores in $\{0,1,\ldots,10\}$ for each of the
$N_{\text{LLM}}$ initial candidates. These scores are treated as
observations and added to $\mathcal{D}_0$. 

\paragraph{Question generation and answer parsing.}
Given the contrasting pair $(\phi_t^{\text{best}}, \phi_t^{\text{query}})$,
a templated prompt asks the LLM to identify the most discriminative
attribute between the two formulas (object identity, time window, ordering,
or safety constraint) and to verbalize a single clarification question
targeting that attribute. The user's free-form response, the question, and
the contrasting pair are then passed to a second LLM call that updates the
candidate pool, adding or removing candidates as warranted, and emits
fresh integer utility scores for every candidate that was affected by the
response. The new scores are appended to $\mathcal{D}_t$ and the GP is
refit.

\subsection{Grammar VAE}
\label{app:vae}

\paragraph{Grammar and AST encoding.}
We instantiate a context-free grammar over the temporal and Boolean
operators used in our task family. The non-terminal set is
\[
\{
S,\,
\text{\textsc{TempUnaryOp}},\,
\text{\textsc{TempBinaryOp}},\,
\text{\textsc{LogUnaryOp}},\,
\text{\textsc{LogBinaryOp}},\,
\text{\textsc{Ts}},\,
\text{\textsc{Te}},\,
\text{\textsc{AP}}
\},
\]
with $S$ as the start symbol. Temporal terminals are $\{G,F,U\}$, logical
terminals are $\{\neg,\wedge,\vee,\Rightarrow\}$, and time-interval bounds
are drawn from $K$ discretization clusters, with one \textsc{Ts} rule and one
\textsc{Te} rule per cluster. The \textsc{AP} non-terminal expands into one
production per grounded predicate; for STL domains this includes both
\texttt{enter(}$a$\texttt{)} and \texttt{not\_enter(}$a$\texttt{)} for each
area $a$, while LTL grammars use bare proposition names. 

Each formula is first parsed into a derivation-rule sequence in pre-order
using a domain-specific parser, then encoded as a one-hot matrix of shape
$(L, R)$ with a single $1$ at the active rule id in each position, where
$R$ is the grammar's rule count. Sequences are right-padded with zero rows
to a fixed length $L_{\max}=100$, which comfortably exceeds the longest
formula in our training corpus.

\paragraph{Model architecture.}
The grammar VAE follows the design of Kusner et al. The encoder is a BiLSTM (two layers, hidden size $H/2$
per direction). The decoder is a two-layer GRU with hidden size $H$ followed by a
linear projection to rule logits. Crucially, decoding is grammar-masked at
every step: we maintain a stack of unresolved non-terminals (initialized
with the start symbol), retrieve the per-non-terminal binary mask of
admissible rules, subtract a large negative constant from disallowed
logits, and push the right-hand side of the chosen rule onto the stack in
reverse order. This guarantees that every decoded sample, including samples
drawn from arbitrary points in the latent space, corresponds to a
syntactically valid STL or LTL formula. We use $d=16$, $H=256$, $L_{\max}=100$.

\paragraph{Training corpus.}
For each domain we generate $100000$ random formulas using the
domain-specific generator: STL templates draw from the four families used
in the main experiments with bounds $\text{max\_goals}=5$,
$\text{max\_obstacles}=3$, $\text{max\_dependencies}=3$, and uniform
template weights; LTL training data is sampled on a grid of
$(n_{\text{goals}}, \text{template})$ pairs with
$n_{\text{goals}}\in[5,8]$ and templates
$\{\text{visit\_all},\text{sequential},\text{partial\_order}\}$, allocating
an equal number of samples per cell and setting
$n_{\text{obstacles}}=\lceil n_{\text{goals}}/2 \rceil$. The corpus is
split $80/20$ into training and validation.

\paragraph{Training objective and schedule.}
We optimize the standard ELBO
\[
\mathcal{L} = \mathbb{E}_{q_\psi(z\mid\phi)}\!\bigl[\log p_\psi(\phi\mid z)\bigr]
- \beta(t)\,\mathrm{KL}\bigl(q_\psi(z\mid\phi)\,\|\,p(z)\bigr),
\]
with cross-entropy reconstruction loss over the rule sequence and a
standard isotropic Gaussian prior $p(z)=\mathcal{N}(0,I_d)$. We anneal the
KL weight linearly from $0$ to $\beta=1$ over the first $10$ epochs and
optionally apply a per-dimension free-bits floor (disabled in the reported
runs). Training uses Adam with learning rate $10^{-3}$, batch size $32$,
$150$ epochs, gradient-norm clipping at $1.0$, and a
\texttt{ReduceLROnPlateau} scheduler (patience $5$, factor $0.5$).
Decoding during training is in free-running mode, the GRU consumes
softmaxed predictions from the previous step rather than ground-truth
tokens, to better match inference-time behavior of the grammar-masked
decoder.

\paragraph{Checkpoint selection.}
At the end of each epoch we evaluate the validation set by greedy
grammar-masked decoding and record the token-exact reconstruction rate
(fraction of formulas whose decoded rule sequence matches the ground-truth
sequence at every position). The best checkpoint is the one with the
highest token-exact rate, breaking ties by lower validation ELBO. Each VAE
is trained on a single NVIDIA RTX 5090 GPU.

\section{Evaluation and Deployment Setup Details}
\label{app:eval-setup}

 We first describe the shared test-case generation procedure used across all settings, and then distinguish the large-scale simulation evaluation from the simpler real-world deployment profile. The key distinction is that the simulation evaluation is designed for systematic quantitative comparison across domains, LLMs, methods, templates, and difficulty levels, whereas the real-world deployment uses a smaller set of feasible, physically executable tasks grounded in the deployed environment.

\subsection{Setup Overview}

All experiments use a two-stage pipeline. First, test cases are generated offline and stored in a static JSON manifest. Each manifest entry records a stable cell key, the specification-generation parameters, the ground-truth specification $\phi^\star$, and the ambiguous initial instruction $I_0$. Second, each method is evaluated on the same $(\phi^\star, I_0)$ pair for a given cell. This ensures that differences in success rate or clarification rounds are attributable to the clarification mechanism rather than to variation in the sampled task or instruction wording.

\subsection{Shared Test-Case Generation}
\label{app:scenario-generation}

\paragraph{Template families.}
For each cell, the generator samples a ground-truth task specification $\phi^\star$ from one of four template families:
\begin{itemize}
    \item \textbf{Single-goal} (\texttt{tt0}): a single reachability objective $\mathbf{F}_{[a,b]}p_{\text{goal}}$ with an avoidance constraint $\mathbf{G}_{[0,H]}\neg p_{\text{obs}}$.
    \item \textbf{Multi-goal} (\texttt{tt1}): a conjunction $\bigwedge_i \mathbf{F}_{[a_i,b_i]}p_i$ with one or more avoidance constraints and no ordering requirement.
    \item \textbf{Sequential} (\texttt{tt2}): a nested reachability formula $\mathbf{F}(p_1 \wedge \mathbf{F}(p_2 \wedge \cdots))$, enforcing a total order over the goals.
    \item \textbf{Partial-order} (\texttt{tt3}): a multi-goal specification with precedence pairs $(p_i \prec p_j)$ and up to $u$ bounded-until clauses $\neg p_r\,\mathbf{U}_{[a,b]}\,p_p$, meaning that region $r$ should not be entered until $p$ has been reached.
\end{itemize}

\paragraph{Ground-truth sampling and feasibility filtering.}
Each cell fixes the template type and its count parameters, such as the number of goals, obstacles, sequential steps, precedence constraints, and until clauses. Goal and obstacle atoms are sampled from the grounded proposition set $\Pi_{\mathcal{G}}$ of the corresponding domain. For STL, bounded intervals $[a,b]$ are sampled within a horizon of $H=60$ seconds, conditioned on $a<b$, except in the real-world deployment profile described below, where intervals are fixed to the full horizon.

After sampling, each candidate $\phi^\star$ is passed to the same downstream planner used during evaluation. STL specifications are checked using the MILP/gradient-based planner, while LTL specifications are checked through the automaton-product planner. If no satisfying trajectory is found within the allowed feasibility budget, the specification is rejected and resampled. This filtering step ensures that failures during evaluation are due to clarification or inference errors rather than infeasible ground-truth tasks.

\subsection{Systematic Ambiguity Injection}

The ambiguous instruction $I_0$ is generated from $\phi^\star$ using a deterministic ambiguity plan. The plan removes or weakens information along four axes before prompting the instruction-writing LLM to produce a one-sentence underspecified instruction. For an STL specification with $g$ goals, $o$ obstacles, $T$ ordering or until dependencies, and $I$ bounded intervals, the ambiguity categories are:

\begin{enumerate}
    \item[\textbf{C1}] \emph{Referential vagueness.} $\lceil g/2 \rceil$ goals are referred to only by category, attribute, or role. In Panda, this may produce references such as ``the red block'' rather than the exact object name. In School, a goal is eligible only when its semantic category has at least two instances in the scene, so the instruction may say ``a trash can'' rather than \texttt{trash\_3}.
    
    \item[\textbf{C2}] \emph{Temporal-dependency omission.} $\lceil T/2 \rceil$ ordering pairs or until clauses are omitted from the surface form. Sequential goals may be mentioned without explicit ``first'', ``then'', or ``before'' language, and conditional avoidance constraints may be omitted or phrased as permanent avoidance.
    
    \item[\textbf{C3}] \emph{Goal/obstacle omission.} Among the goals and obstacles not already affected by C1 or C2, approximately half are omitted entirely from the utterance.
    
    \item[\textbf{C4}] \emph{Time-interval omission.} For STL only, $\lceil I/2 \rceil$ bounded $\mathbf{F}$, $\mathbf{G}$, or $\mathbf{U}$ windows are dropped. The relevant action or region is still mentioned, but the seconds-level deadline or duration is not. This category is not used for LTL because LTL has no metric time intervals.
\end{enumerate}

The categories are applied in order, so that an item omitted by an earlier category is not double-counted by a later one. Items that remain after ambiguity injection must be stated explicitly, with time intervals expressed in plain seconds rather than formula notation. The ambiguity plan and the resulting instruction are cached in the manifest so that all methods and base LLMs receive the same initial instruction for each cell.

\subsection{Simulation Evaluation Setup}
\label{app:simulation-setup}

The simulation evaluation consists of two production sweeps used to populate the main quantitative results: the Panda STL sweep and the City LTL sweep. These sweeps are designed to stress-test the methods across domains, templates, difficulty levels, LLM backbones, and random seeds.

\paragraph{Simulation complexity levels.}
The simulation sweeps use larger template-aware complexity grids. Panda STL varies the number of goals from one to five, with one obstacle and one until clause. City LTL varies the number of goals from five to eight, with obstacle and constraint counts scaled with the number of goals.  We keep complexity template-aware rather than treating goal count as the only difficulty measure, because sequential and partial-order specifications impose additional structural constraints on the planner.

\subsection{Real-World Deployment Setup}
\label{app:realworld-setup}

The real-world deployment uses the same overall task-generation and clarification framework, but with a smaller and more physically constrained task profile. Unlike the simulation sweeps, which are intended to cover broad systematic variation, the deployment setup prioritizes reliable grounding, feasible planning, and executable robot behavior in the reconstructed environment.

\paragraph{Deployment task profile.}
The deployment tasks use the same template families but operate in a simple regime. Within each template family, the main count parameter is rotated deterministically over $\{1,2,3\}$: multi-goal tasks vary the number of goals, sequential tasks vary the number of ordered steps, and partial-order tasks vary the number of precedence constraints while holding one until clause fixed. Single-goal tasks have no additional count parameter. Each cell also fixes one obstacle. This produces specifications with roughly two to five grounded propositions per cell, making the tasks suitable for real-world execution while still requiring clarification.

\paragraph{Deployment feasibility check.}
Each sampled deployment task is checked with the downstream planner before execution. The planner uses a 60\,s horizon and a nominal robot speed of 3\,m\,s$^{-1}$. If no satisfying trajectory is found within the feasibility budget, the seed is re-rolled up to ten times before the cell is dropped. Thus, the deployment benchmark contains only tasks that are feasible under the planner and environment map.

\paragraph{Single-time-interval encoding.}
For the deployment profile, the grammar VAE is trained with a single time cluster. As a result, every $\mathbf{F}$, $\mathbf{G}$, and $\mathbf{U}$ interval in the ground-truth specification is fixed to $[0,\infty]$, corresponding to the full 60\,s run. Tight metric windows are used only in the larger simulation tiers, where the VAE checkpoint is trained with multiple time clusters. This keeps the deployment latent space well-conditioned while preserving temporal expressiveness in the simulation stress tests.

\paragraph{Robot execution.}
After clarification, the inferred task specification is grounded to the deployment scene and passed to the formal planner. The planner synthesizes a trajectory that satisfies the inferred STL specification, and the resulting trajectory is executed by the robot in the real-world environment. In contrast to the simulation sweeps, the deployment setting emphasizes whether the full language-to-specification-to-planning pipeline can produce a physically executable behavior.

\subsection{Human Proxy and Approval}
\label{app:human-proxy}

\paragraph{Proxy conditioning.}
For each test case, the proxy is conditioned on a short STL/LTL semantic primer, the raw ground-truth specification $\phi^\star$, the scene context exposed to the helper LLM, and the list of ground-truth goals and obstacles. The proxy is explicitly instructed not to reveal formulas, predicate names, option labels, or interval-bracket notation. When time constraints are present, they must be expressed in plain natural language, such as ``within 24 seconds'' or ``for the first 18 seconds.'' The proxy answers only the first detected question in a single short sentence and does not volunteer information beyond what was asked.

\paragraph{Proxy/helper decoupling.}
The proxy is fixed to GPT-5.4 with medium reasoning effort, independent of the helper LLM being evaluated. This decouples the simulated user's behavior from the model under test and isolates performance differences along the method axis.

\paragraph{Approval and termination.}
Approval is separated from the proxy. At each clarification round, the candidate trajectory is checked against $\phi^\star$ using the domain-appropriate verifier: sampled STL robustness for Panda, STL trajectory satisfaction against the School area map, and LTL satisfaction on the discrete product automaton for City. A method is credited with success when the generated trajectory satisfies the ground-truth specification and the interaction terminates at that round. The query budget is capped at $B=10$; runs that reach the cap without satisfaction are recorded as failures.
\section{Safe Autonomous Navigation}

A core requirement of the robot deployment is to execute the inferred task specification safely in cluttered real-world environments. We implement a single-agent safe navigation stack on the Unitree Go2 quadruped robot. The stack consists of three layers: a model predictive controller (MPC), a control barrier function (CBF) safety filter, and the robot's built-in low-level locomotion controller. At each control cycle, the robot state, including position, heading, and velocity, is estimated by fusing odometry, IMU, and LiDAR measurements. The estimated state is then passed to the planning and control stack.

\subsection*{Model Predictive Control with Safety Certificates}

\paragraph{MPC planner.}
The MPC planner generates a locally feasible trajectory by solving a nonlinear program over a receding horizon of \(T\) steps. We model the robot using a second-order unicycle model, where the state is
\[
x_k = [p_x, p_y, \theta, v, \omega]^\top,
\]
and the control input is
\[
u_k = [a, \alpha]^\top,
\]
with \(a\) and \(\alpha\) denoting linear and angular acceleration, respectively. At each timestep, MPC solves
\begin{equation}
    \min_{\{u_k\}_{k=0}^{T-1}}
    \sum_{k=0}^{T-1} l(x_k,u_k) + l_f(x_T)
    \quad
    \text{s.t.}
    \quad
    \begin{cases}
        x_{k+1} = f(x_k,u_k), \\
        h^{(m)}(x_k) \geq 0, \quad \forall m, \\
        u_k \in \mathcal{U},
    \end{cases}
\end{equation}
where \(l(\cdot)\) and \(l_f(\cdot)\) are the running and terminal tracking costs, \(h^{(m)}(x_k)\) encodes the signed safety margin with respect to obstacle \(m\), and \(\mathcal{U}\) denotes the admissible input set. The safe set is defined as
\[
\mathcal{C}^{(m)} = \{x \mid h^{(m)}(x) \geq 0\}.
\]
The nonlinear program is solved online, and only the first control input \(u_0^*\) is applied before the optimization is repeated at the next control cycle.

\paragraph{CBF safety filter.}
Although MPC plans over a finite horizon, model mismatch, localization error, and timing delays can still lead to unsafe commands. Therefore, before execution, the MPC command is passed through a CBF safety filter. The CBF layer operates on a first-order unicycle model and solves a quadratic program:
\begin{equation}
    \min_{u \in \mathcal{U}} \|u - u_0^*\|^2
    \quad
    \text{s.t.}
    \quad
    \dot{h}^{(m)}(x,u) + \gamma h^{(m)}(x) \geq 0,
    \quad \forall m,
\end{equation}
where \(h^{(m)}\) is the barrier function for obstacle \(m\), and \(\gamma > 0\) controls how aggressively the robot maintains distance from the safety boundary. This QP minimally modifies the MPC command while enforcing instantaneous safety constraints. In this way, MPC provides long-horizon goal-directed planning, while the CBF layer provides a local safety certificate at execution time.

\paragraph{Low-level locomotion controller.}
The filtered command is converted into velocity commands, including longitudinal velocity and yaw rate, and sent to the Unitree Go2's built-in low-level controller. The onboard controller tracks these commands through whole-body locomotion control and generates the corresponding joint-level actions for the quadruped.

\section{Robotic Platform and Perception Stack}

\subsection{Robotic Platform}

We deploy the system on the Unitree Go2 quadruped robot. The Go2 provides an agile mobile platform for autonomous inspection in indoor and industrial environments. Its onboard computation and sensor interfaces allow integration with LiDAR, inertial sensing, odometry, and RGB-D perception modules.

\subsection{Sensor Suite}

The deployed robot uses the following sensing modalities:

\begin{itemize}
    \item \textbf{LiDAR.} We support two LiDAR configurations depending on the deployment setting. The basic configuration uses the Unitree 4D LiDAR L1 for lightweight indoor navigation. The advanced configuration uses the Hesai XT16, a 16-channel 3D LiDAR that provides denser point clouds for more complex environments. LiDAR measurements are used for obstacle detection, mapping, and localization.

    \item \textbf{Odometry.} Legged odometry is computed from the robot's proprioceptive measurements and kinematic model. It provides short-term motion estimates that improve local consistency during state estimation.

    \item \textbf{Inertial Measurement Unit.} The onboard IMU provides high-frequency linear acceleration and angular velocity measurements. These measurements are fused with odometry and LiDAR observations for robust pose estimation.

    \item \textbf{RGB-D Camera.} An Intel RealSense camera provides RGB-D observations for semantic perception. The RGB-D stream is used to support object-level understanding of the environment and to ground high-level task specifications to perceived scene elements.
\end{itemize}

\subsection{Mapping}

To use rich 3D sensing with standard 2D navigation tools, we convert LiDAR observations into a planar representation. Specifically, the 3D point cloud is filtered by height to remove irrelevant points, such as the floor and overhead structures. The remaining points are projected into a 2D laser scan representation, which is then used to construct a 2D occupancy grid map of the environment. This map provides the geometric prior used by localization and navigation.

\subsection{Localization}

Robot localization is performed using Adaptive Monte Carlo Localization (AMCL). AMCL estimates the robot pose on the 2D occupancy grid map by combining laser scan observations with odometry inputs. The method maintains a weighted set of particles representing possible robot poses and updates these particles through Bayesian filtering. Adaptive resampling adjusts the number of particles online, improving robustness and computational efficiency during deployment.

The resulting pose estimate is used by the MPC-CBF navigation stack to compute safe velocity commands in the map frame.
\section{Human User Study}
\label{app:user-study}

\subsection{Participants and Design}

We recruited participants through Prolific and retained $N{=}30$
after excluding incomplete submissions. All participants were
at least 18 years old and provided informed consent. The study
was approved by the IRB as exempt research. Participation took
approximately 30 minutes, and each participant received US\$6.

We used a within-subject design with six episodes per participant:
two each for \textit{BAL}, \textit{LM}, and \textit{LM+P}.
Method order was counterbalanced using the cyclic Latin-square
sequences $ABCABC$, $BCABCA$, and $CABCAB$, where $A$, $B$,
and $C$ denote the three methods. Task order was randomized
independently.

\subsection{Task Setup and Participant Role}

The study followed the same task-generation and clarification
protocol as the simulation evaluation, with human participants
replacing the LLM proxy as clarification respondents. Each case
paired a predefined ground-truth task specification with an
ambiguous initial instruction generated offline using the
systematic ambiguity-injection procedure described in
Appendix~\ref{app:scenario-generation}. Participants neither
defined the intended task nor composed the initial instruction.

Participants viewed the ground-truth task on a city map,
including required destinations, forbidden locations, and ordering
constraints. The agent received only the environment context and
the generated ambiguous instruction. Participants answered
subsequent clarification questions according to the displayed
task. As in simulation, they were instructed to answer only the
question asked without volunteering additional information.

The six cases included two each from the multi-goal, sequential,
and partial-order template families. Each case contained four
targets and one forbidden location, with until clauses disabled.
Compared with simulation, the study used simpler tasks and a
reduced budget of five clarification rounds per episode.

\subsection{Procedure and Interface}
\label{app:user-study-interface}

Participants completed a brief tutorial and a comprehension check
before beginning the six episodes. In each episode, they inspected
the assigned task and answered the agent's questions in free-form
natural language.

We used the same \emph{human-approval} stopping condition as in
the main simulation experiments, implemented through automated
ground-truth verification. At each round, the proposed plan was
checked against the ground-truth task specification, and the
episode terminated once all requirements were satisfied or the
five-round budget was exhausted. Participants were not asked
to evaluate or approve the proposed plan; their role was limited
to answering clarification questions.

\begin{figure}[t]
    \centering
    \includegraphics[width=\linewidth]{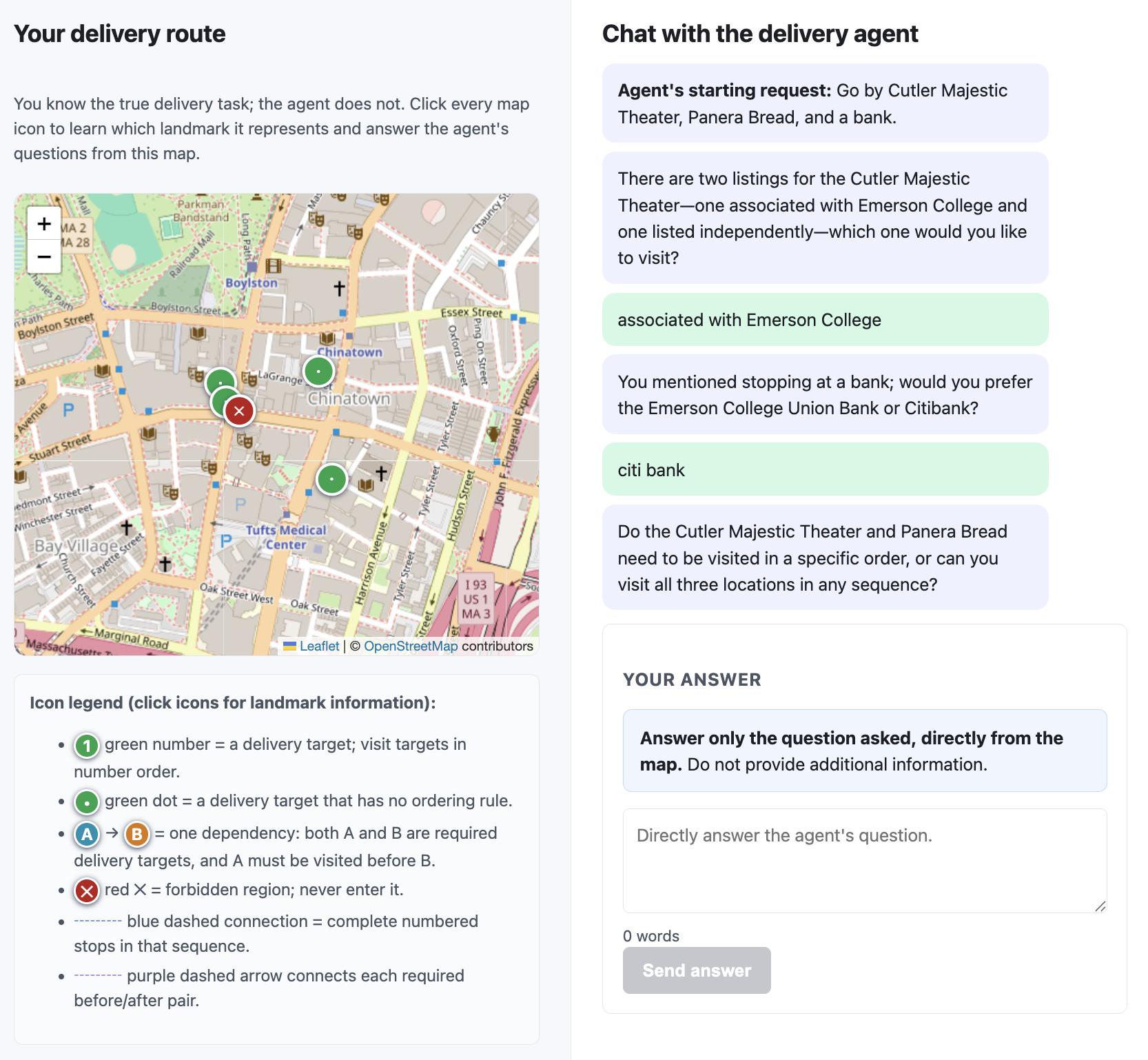}
    \caption{City user-study interface. Participants view an
    assigned ground-truth task on the map and answer clarification
    questions through the chat panel. The agent receives a
    systematically generated ambiguous instruction without access
    to the ground-truth specification.}
    \label{fig:user-study-ui}
\end{figure}

Figure~\ref{fig:user-study-ui} shows the interface shared by all
three methods. The left panel displayed an interactive city map
with targets, forbidden locations, and ordering constraints.
Clicking a landmark revealed its name and description; formal
LTL formulas were not shown. The right panel displayed the
clarification dialogue and response box. When available, the
generated plan was overlaid on the map.

\subsection{Outcome Analysis}

The primary outcomes were task success rate and clarification
rounds. Success indicated that the generated plan satisfied the
ground-truth specification within the five-round budget.
Each outcome was averaged over the two episodes assigned to
each method for each participant. The main paper reports
means and standard errors across the $N{=}30$ participant-level
averages. Method effects were assessed using one-way
repeated-measures ANOVA with method as the within-subject factor.

\section{Prompts}
\label{app:prompt}

This appendix complements the Experiments section by documenting how the
three methods (\textbf{BAL}, \textbf{LLM}, and \textbf{LLM+P}) are wired
to the same LLM backends, the same simulated user, and the same
per-task context. We then list, verbatim, every prompt template invoked
during the simulated sweeps reported in the main paper.

\subsection{System design and role of LLMs}
\label{app:eval-roles}

Each test case is parameterized by a tuple
$(\mathcal{G}, \phi^\star, I_0, \pi_{\text{LLM}}, \pi_{\text{proxy}})$:
a grounded scene $\mathcal{G}$, a hidden ground-truth specification
$\phi^\star$, an ambiguous instruction $I_0$, a base LLM $\pi_{\text{LLM}}$
under test, and a proxy LLM $\pi_{\text{proxy}}$ that simulates the user.
$\pi_{\text{proxy}}$ is fixed to GPT-5.4 across the sweeps and is the
\emph{only} module that ever sees $\phi^\star$. The base LLM is swapped
across the six models while the rest of the
pipeline is held constant.

\paragraph{LLM roles in BAL.} A single base LLM is invoked in four
distinct roles. (R1)~\textsc{Candidate generator + prior scorer}:
called once on $I_0$ to seed the candidate pool $\mathcal{C}_0$ with
$N{=}20$ STL/LTL formulas plus a 1--10 prior score per candidate, and
called again at every clarification round to regenerate $\mathcal{C}_t$
conditioned on the full Q/A history.
(R2)~\textsc{Clarification verbalizer}: called once per round to turn
the latent contrast $(\phi_t^{\text{best}}, \phi_t^{\text{query}})$ into
a single natural-language question. The Bayesian active learning loop itself
(latent encoding, GP fit, mutual-information acquisition, MILP/Spot
planning) does not call the LLM.

\paragraph{LLM roles in the baselines.} The two ablation baselines
deliberately replace some of these calls with the LLM acting as its own
clarification driver. \textbf{LLM} keeps two roles: (R2$'$)~a free-form
clarification asker that has \emph{no} candidate contrast to anchor on,
and (R$_{\text{plan}}$)~a direct planner that emits the executable
artifact (School/Factory: 2-D waypoints; Panda: 3-D EE waypoints;
City: ordered landmark list). \textbf{LLM+P} also keeps R2$'$ but
replaces R$_{\text{plan}}$ with R$_{\text{spec}}$: the LLM emits a
\emph{single} STL/LTL formula, which is then passed to the same
downstream planner used by BAL (MILP for School/Factory, gradient
optimizer for Panda, Spot product for City). LLM+P therefore isolates
the contribution of the Bayesian uncertainty model from the
contribution of the formal planner.

\paragraph{LLM roles in the simulated user.} The proxy LLM serves two
roles, both conditioned on the hidden $\phi^\star$. (P1)~\textsc{Initial
instruction generator}: generates $I_0$ from $\phi^\star$ via the
\emph{systematic ambiguity} prompt that hides 0--2 axes per category
across (referential, temporal-dependency, goal/obstacle, time-interval)
to produce ${\sim}50\%$ ambiguity per axis. (P2)~\textsc{Clarification
answerer}: answers each clarification question in 1--2 plain sentences
without leaking the formula. 

\subsection{Context parity across methods}
\label{app:context-parity}

For every test case, the three methods receive the exact same inputs at
every interaction point. Concretely:
\begin{itemize}\itemsep1pt
  \item \textbf{Same scene block.} A single \texttt{\{context\}} string
  is built once per scenario. It is interpolated into every BAL, LLM, LLM+P, and proxy prompt, no
  method sees extra environment detail.
  \item \textbf{Same predicate vocabulary.} Goals and obstacles are
  drawn from the same atomic-proposition set $\Pi_{\mathcal{G}}$ in
  every prompt. The same block is included in every
  spec-emitting prompt (BAL R1, LLM+P R$_{\text{spec}}$, and the
  diagnostic-only oracle rater); it is excluded from the
  free-form-waypoint paths so the LLM baseline is not forced into STL.
  \item \textbf{Same instruction $I_0$ and proxy.} Both are sampled
  once per test case from a fixed seed and are reused verbatim across
  the three methods, so any difference in performance reflects
  algorithmic choice rather than environment variance.
  \item \textbf{Same budget and stopping rule.} $B{=}10$ clarification
  rounds, with early stop on proxy approval, applies uniformly. 
  \item \textbf{Same model configuration.} Both $\pi_{\text{LLM}}$ and
  $\pi_{\text{proxy}}$ use the same base model and reasoning effort.
\end{itemize}

The remaining difference between methods is therefore the prompt that
each LLM call uses. The next subsections enumerate those prompts.

\subsection{Shared simulated-user prompts}
\label{app:proxy-prompts}

These prompts drive the proxy LLM and are identical across BAL, LLM,
and LLM+P runs.

\begin{promptbox}{P1 — Systematic ambiguous-instruction generator}
\footnotesize\ttfamily
Background: Signal Temporal Logic (STL) and templates\\
\{stl\_knowledge\}\\[2pt]
Ground truth intent for this scenario (for evaluation; do not reveal literally):\\
- Template: \{template\_name\}\\
- Goals: \{goals\}\\
- Obstacles to avoid: \{obstacles\}\\
- Raw STL formula: \{formula\}\\[4pt]
Task: Generate ONE short, natural-sounding navigation instruction for the ground-truth task above. Apply EXACTLY the ambiguity plan listed below — no more, no less. The target is $\sim$50\% ambiguity per axis: about half of each axis is left unclear.\\[4pt]
CATEGORY 1 — REFERENTIAL AMBIGUITY: \{cat1\_text\}\\
CATEGORY 2 — TEMPORAL DEPENDENCY OMISSION: \{cat2\_text\}\\
CATEGORY 3 — GOAL / OBSTACLE UNDERSPECIFICATION: \{cat3\_text\}\\
CATEGORY 4 — TIME-INTERVAL OMISSION: \{cat4\_text\}\\[4pt]
ALL goals, obstacles, and constraints NOT listed above MUST appear: goals not in Cat.~1/3 by exact name; obstacles not in Cat.~3 stated as avoidance; dependencies/Until not in Cat.~2 stated explicitly; intervals not in Cat.~4 stated in precise seconds.\\[2pt]
HARD RULES: 1.~ONE short utterance (1--2 sentences). 2.~No formal logic or option labels. 3.~Preserve every Until clause unless explicitly in Cat.~2. 4.~Output only the instruction text.
\end{promptbox}

\begin{promptbox}{P2 — Proxy clarification answerer}
\footnotesize\ttfamily
Background — Signal Temporal Logic (STL) for robot navigation:\\
\{stl\_knowledge\}\\[2pt]
Your true intent is specified by the following ground-truth STL formula:\\
\{gt\_formula\}\\[2pt]
All STL interval values in the ground truth are REAL SECONDS in [0, 60]; 60 and inf mean the same thing. When the GT contains a TIGHT window, translate it directly into natural-language seconds:\\
\quad F[10,30](enter(X))    $\rightarrow$ "reach X between 10 and 30 seconds in"\\
\quad G[0,18](not\_enter(Y)) $\rightarrow$ "keep clear of Y for the first 18 seconds"\\
\quad G[12,48](not\_enter(Y)) $\rightarrow$ "keep clear of Y between 12 and 48 seconds"\\[4pt]
You are simulating a user answering a clarification question. Based ONLY on your true intent, answer the following in 1--2 short natural sentences, as a real user would. If the text contains multiple questions, answer ONLY the first. Do not mention formulas, option labels, or STL.\\[2pt]
Clarification question: \{question\}\\
Your answer:
\end{promptbox}

\subsection{BAL prompts}
\label{app:bal-prompts}

These prompts cover R1 and R2 of the BAL loop. The School (\emph{robot-dog
navigation}) and Panda (\emph{tabletop manipulation}) packs share the
same structure; only the agent description, predicate semantics, and
ICL exemplars differ. The LTL pack is structurally identical but uses
prefix-form formulas and the 4 LTL templates (visit\_one, visit\_all,
sequential, partial\_order).


\begin{promptbox}{R1 — Candidate generation + prior with NL feedback}
\footnotesize\ttfamily
You are an expert in Signal Temporal Logic (STL) for robot dog navigation in a building. The robot has limited sensing range therefore must enter the target object to inspect it.\\[2pt]
Generate \{n\} different valid STL interpretations of the instruction. For EACH interpretation, rate (1--10, 10 = perfect match) how well it captures the user's intent.\\[2pt]
Respond with a JSON array containing EXACTLY \{n\} elements, each with EXACTLY three fields:\\
- \texttt{"stl\_formula"}: full formula string\\
- \texttt{"template\_type"}: int 0|1|2|3\\
- \texttt{"prior\_score"}: int 1--10\\[2pt]
[STL\_FULL\_FORMULA\_EXAMPLES — canonical shapes for templates 0/1/2/3]\\
\{context\}\\[2pt]
Original task instruction:\\\{instruction\}\\[2pt]
Clarification history so far (oldest $\rightarrow$ newest):\\\{qa\_history\}\\[2pt]
Your job is to PROGRESSIVELY NARROW the space of plausible STL interpretations as more clarifications arrive:\\
- Every candidate MUST be consistent with every prior answer.\\
- Commit to details the human has confirmed (specific goal, ordering, avoidance).\\
- Focus remaining diversity on STILL-ambiguous dimensions.\\
- If history pins down the true intent, concentrate prior\_score on the matching formula.\\
- If the human said "no" to avoidance/ordering/multi-goal, do NOT re-add those clauses.\\[2pt]
[STL\_FULL\_FORMULA\_EXAMPLES + INTENT\_DIVERSITY blocks repeated]\\[2pt]
Generate \{n\} different valid STL interpretations; same JSON schema as R1a.\\
Pay attention to timing cues from BOTH the original instruction and the clarification history. Copy concrete second numbers into ts/te and preserve them across regenerations.
\end{promptbox}

\begin{promptbox}{R2 — Clarification-question verbalizer for BAL}
\footnotesize\ttfamily
You are helping a robot dog understand a human's navigation instruction. You are shown the original instruction, the full clarification history, and two specific candidate STL interpretations (Option A and Option B). The two options are ONE example of a dimension that is currently ambiguous — use them as inspiration but write a HIGH-LEVEL OPEN-ENDED question that reveals NEW intent.\\[2pt]
Guiding principles:\\
- Read the clarification history first. Do NOT re-ask resolved dimensions.\\
- Prefer OPEN-ENDED over forced binary choices.\\
- Consider every axis: which one, when (deadline), not before when (start), for how long (duration), in what sequence, all vs.\ any, anything to avoid.\\
- If options differ on a TIME-WINDOW dimension, ASK in concrete seconds.\\
- Never say "Option A"/"Option B"/"STL"/"formula". Do not quote $\mathtt{enter}(\cdot)$ predicates. 1 sentence.\\[2pt]
[10 ICL exemplars covering goal/obstacle/order/duration/until]\\[2pt]
\{context\}\\
Original instruction: "\{instruction\}"\\
Clarification history so far: \{qa\_history\}\\
Option A: \{stl\_a\}\\
Option B: \{stl\_b\}\\[2pt]
Respond with ONLY the question.
\end{promptbox}

\subsection{LLM baseline prompts (clarification-only)}
\label{app:llm-prompts}

\textbf{LLM} keeps the same \texttt{\{context\}} block as BAL but
exposes the model directly to the trajectory output space. The
clarification asker is given no STL contrast — it must invent its own
informative question every round. The terminator is the same
\texttt{DONE}-sentinel mechanism BAL uses for early stopping.

\begin{promptbox}{R2 — Free-form clarification asker}
\footnotesize\ttfamily
Context (environment and areas):\\\{context\}\\
User instruction: \{instruction\}\\
So far you have asked the following clarification questions and received answers: \{qa\_block\}\\[2pt]
You are deciding between two actions this round:\\
\quad (a) Ask exactly ONE more short clarification question, OR\\
\quad (b) Stop because you already have enough information.\\[2pt]
Useful topics: which goal area, area arrival deadline, goal start window (earliest start), how long to avoid an area, visit order between goals, areas to avoid. Prefer concrete second numbers over vague "soon/later".\\[2pt]
Output rules (pick exactly one format):\\
\quad (a) one short question ending in "?". No compound questions, no "and"/"also"/"then", no bullets, no preamble.\\
\quad (b) the single literal token DONE on its own line, no other text.
\end{promptbox}

\begin{promptbox}{R$_{\text{plan}}$ — clarification $\rightarrow$ waypoints}
\footnotesize\ttfamily
You are a robot navigation planner. The user gave an ambiguous instruction; you asked clarification questions and received answers. Now output a single trajectory as a JSON array of waypoints.\\[2pt]
Environment and areas (with bounding boxes):\\\{context\}\\
Original instruction: \{instruction\}\\
Clarification Q\&A:\\\{qa\_block\}\\[2pt]
Requirements:\\
- The first waypoint must be exactly the robot's start position in the context.\\
- Waypoints must lie strictly within the room boundaries described above.\\
- Each waypoint is \{"x": float, "y": float, "t": optional float\}.\\[2pt]
If the requested constraints cannot all be satisfied within the horizon and the robot's speed limit, output the single token INFEASIBLE on its own line. Otherwise output only the JSON array, no other text.
\end{promptbox}

\subsection{LLM+P baseline prompts (clarification $\rightarrow$ formal
spec $\rightarrow$ planner)}
\label{app:llmp-prompts}

\textbf{LLM+P} replaces R$_{\text{plan}}$ with a spec-emitting stage
that re-uses the BAL R1 prompt, but constrained to a single best
candidate ($n{=}1$). It uses the \emph{same} free-form clarification
asker as LLM (R2), so the only differences between LLM and LLM+P at
runtime are: (i)~the output schema (one STL formula vs.~one waypoint
list), and (ii)~that the formula is then handed to the same solver BAL
uses. This isolates the contribution of the formal-spec step from the
contribution of Bayesian uncertainty modelling.

\begin{promptbox}{R$_{\text{spec}}$ — single best STL/LTL formula
under full Q/A history}
\footnotesize\ttfamily
[Same body as BAL R1 — \texttt{\{context\}}, \texttt{\{instruction\}}, \texttt{\{qa\_history\}}, the same FULL\_FORMULA\_EXAMPLES, INTENT\_DIVERSITY, TIMING\_HARD\_RULES, NESTING\_RULES blocks]\\[2pt]
Generate 1 valid STL interpretation. Respond with a JSON array of EXACTLY one element with the standard schema (\texttt{stl\_formula}, \texttt{template\_type}, \texttt{prior\_score}). The downstream planner will lift this formula into a trajectory using the same MILP / gradient / Spot pipeline BAL uses; do not emit waypoints.
\end{promptbox}

\end{document}